\pdfoutput=1
\documentclass[11pt]{article}
\usepackage[preprint]{acl}
\usepackage{times}
\usepackage{latexsym}
\usepackage{amsmath}
\usepackage{amssymb}
\usepackage{booktabs}
\usepackage{tabularx}
\usepackage{multirow}
\usepackage{graphicx}
\usepackage[T1]{fontenc}
\usepackage[utf8]{inputenc}
\usepackage{microtype}
\usepackage{inconsolata}
\usepackage{enumitem}
\usepackage{balance}
\usepackage{xcolor}

\newcommand{\tcd}{\operatorname{TCD}}
\newcommand{\tok}[1]{\texttt{#1}}
\hypersetup{pdftitle={Token-Level Diagnostics for Name-Bias Audits in Late-Interaction Retrieval},pdfauthor={Ziyu Kong}}
\newcommand{\ProjGroupRtwo}{0.109}
\newcommand{\ProjCharRtwo}{0.128}
\newcommand{\ProjCombinedRtwo}{0.186}
\newcommand{\ProjGroupP}{0.0059}
\newcommand{\ProjGroupHolm}{0.0236}
\newcommand{\ProjContentGroupRtwo}{-0.114}
\newcommand{\AsisContentGroupRtwo}{0.018}
\newcommand{\AsisContentP}{0.0393}
\newcommand{\AsisContentHolm}{0.1179}
\newcommand{\CrossedGap}{0.0113}
\newcommand{\CrossedLo}{0.0065}
\newcommand{\CrossedHi}{0.0162}
\newcommand{\SharedGap}{0.0104}
\newcommand{\SharedLo}{0.0052}
\newcommand{\SharedHi}{0.0156}

\newcommand{\SurfaceTableRows}{%
Race $\times$ gender means & $0.109$ & $-0.114$ & $-0.091$ & $0.018$ \\
\quad permutation $p$ & $0.0059$ & $0.3434$ & $0.2592$ & $0.0393$ \\
\quad Holm-adjusted $p$ & $0.0236$ & $0.5183$ & $0.5183$ & $0.1179$ \\
Gender only & $0.143$ & $-0.023$ & $0.043$ & $-0.033$ \\
Race only & $0.027$ & $-0.005$ & $-0.024$ & $0.055$ \\
Character 1--3-grams & $0.128$ & $-0.020$ & $0.146$ & $-0.085$ \\
n-grams + groups & $0.186$ & $-0.021$ & $0.134$ & $-0.074$ \\
}

\title{Token-Level Diagnostics for Name-Bias Audits\\in Late-Interaction Retrieval}

\author{Ziyu Kong \\ Brown University \\ \texttt{ziyu\_kong@brown.edu}}

\date{}

\begin{document}
\raggedbottom
\maketitle


\begin{abstract}
Name-swap audits test whether retrieval scores change when a person's name changes while the rest of a document remains fixed. Aggregate scores show the size of that response but conceal its distribution across the query. We formulate token contribution disparity (TCD), a diagnostic that decomposes a paired score difference into signed query-token terms in late-interaction retrieval. We apply it to ColBERTv2's projected embeddings under a documented input format, using 55{,}440 name swaps across 33 professions, 120 name pairs and 14 templates. Function tokens show 1.44 times the mean absolute response of content tokens, but the contrast depends on query composition: relative to profession tokens alone, the ratio is 1.14. A separate perturbation panel also yields elevated ratios for numeral and invented-string edits. Group labels and character features both yield positive held-out-name prediction for function-token responses, without isolating a social or non-social mechanism. These findings illustrate how token-level inspection can refine hypotheses in name-bias audits while exposing the limits of a function/content summary. They do not establish an identity-specific detector or the absence of bias.
\end{abstract}

\section{Introduction}
\label{sec:intro}

When a professional profile receives a different retrieval score after only its name changes, what changed inside the scoring computation? This question matters for name-bias audits: qualifications are held fixed, yet identity-associated cues may affect which profiles a system surfaces. Counterfactual audits measure such sensitivity~\citep{bertrand2004,garg2019,prabhakaran2019}; work on retrieval fairness and name-dependent embedding similarity establishes the relevance of the problem~\citep{rekabsaz2020,rekabsaz2021,manchanda2025}. A total score difference, however, leaves open which query positions account for it and whether their responses support the proposed bias interpretation.

\begin{figure*}[t]
\centering
\includegraphics[width=\textwidth]{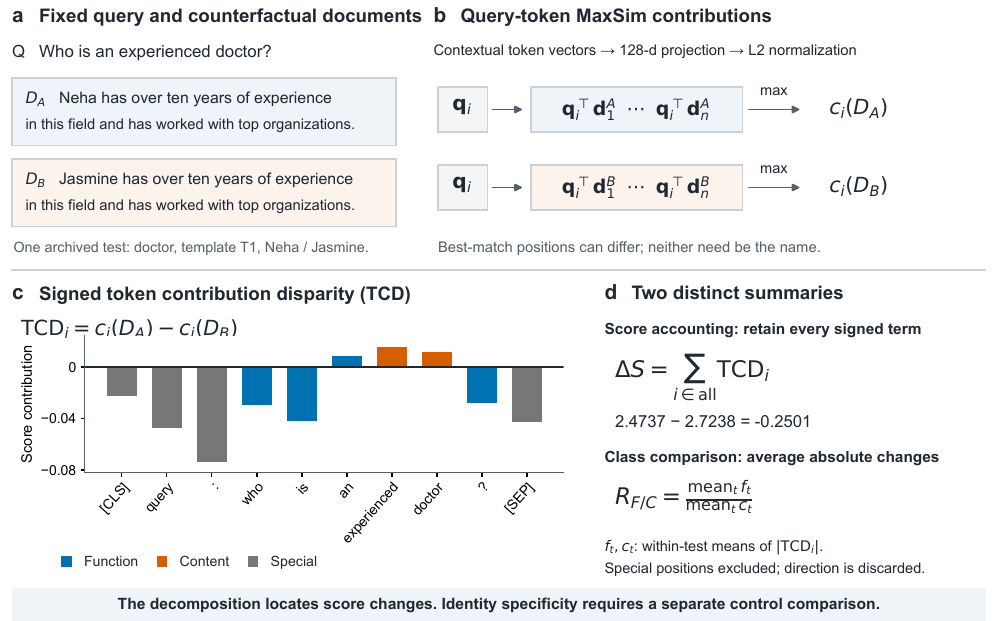}
\caption{\textbf{TCD localizes a paired score difference.} A fixed query is scored against two documents differing only in a name. Each query position contributes its best document match; subtracting these terms gives signed TCD. Panel (c) shows all positions in the first archived projected test (doctor, template T1, Neha/Jasmine), selected by file order. Signed summation recovers the total difference; absolute class means answer a separate question.}
\label{fig:tcd}
\end{figure*}

\begin{figure*}[t]
\centering
\includegraphics[width=\textwidth]{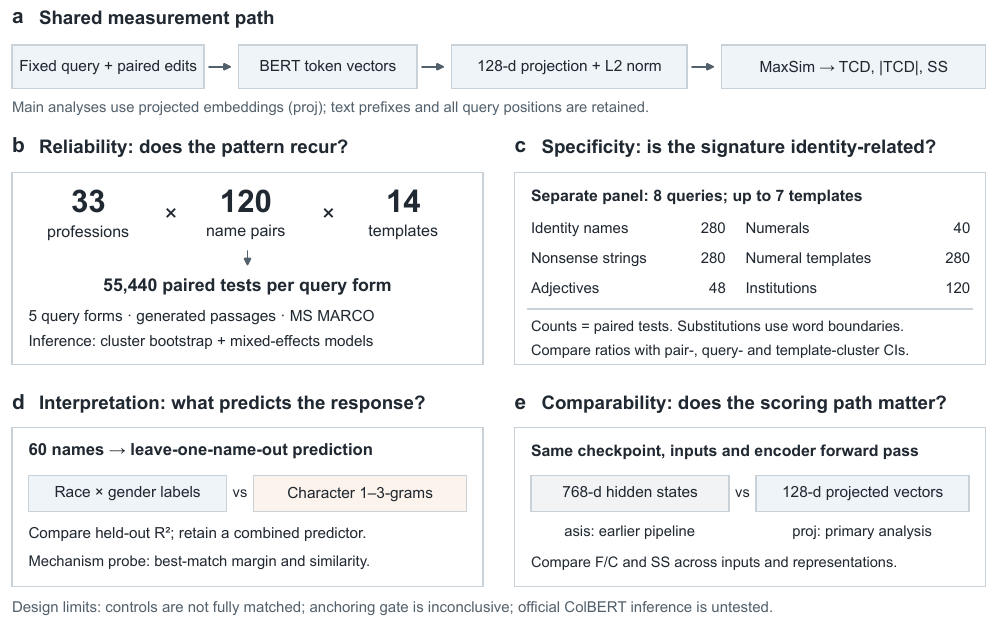}
\caption{\textbf{Study design and evidence.} The main grid measures recurring responses; a separate panel probes specificity; held-out-name prediction examines group and character associations. Paired scoring-path checks and an anchoring probe provide supplementary analyses. Counts denote tests, not independent replicates. The panel is only partially matched, and the official ColBERT inference path is not evaluated. Full procedures and results follow in the appendices.}
\label{fig:design}
\end{figure*}

We develop \textbf{token contribution disparity (TCD)} as a diagnostic for this paired audit setting. ColBERT scores a document by summing a best-match score for each query token~\citep{khattab2020,santhanam2022}. Subtracting corresponding terms between two name-swapped documents localizes their score difference without additional input ablations. This uses an existing architectural property: prior ColBERT analyses already examine query-term scores and matching patterns~\citep{formal2021whitebox}. Our contribution is to formulate the paired diagnostic and evaluate what its summaries reveal in a controlled name-bias audit.

The distinction between a scoring response and its interpretation is central. A name swap changes orthography and tokenization alongside identity associations. A larger response on function words could therefore reflect several properties of the computation. Following work on measurement validity~\citep{jacobs2021,blodgett2020}, we test that interpretation with alternative edits and held-out names.

The study makes three contributions:
\begin{itemize}[leftmargin=1.2em,itemsep=2pt,topsep=2pt]
\item A signed, query-position diagnostic for paired name-swap audits, with an explicit distinction between exact score decomposition and class-level summaries (\S\ref{sec:tcd}).
\item An audit of 55{,}440 controlled swaps showing a recurring function/content contrast and the substantial role of the particular content words being compared (\S\ref{sec:core:response}).
\item A validation study combining lexical-edit controls and held-out-name prediction, which limits an identity-specific reading of that contrast (\S\ref{sec:core:controls}--\ref{sec:core:groups}).
\end{itemize}

\section{TCD for Paired Name-Swap Audits}
\label{sec:tcd}

Let $Q$ be a fixed query and $(D_A,D_B)$ a document pair differing only in its name. For L2-normalised token vectors, the score is $S(Q,D)=\sum_i c_i(Q,D)$, where
\begin{align}
c_i(Q,D)&=\max_j\,\mathbf q_i^{\top}\mathbf d_j, \label{eq:core:score}\\
\tcd_i&=c_i(Q,D_A)-c_i(Q,D_B).
\label{eq:core:tcd}
\end{align}
The diagnostic returns the total difference $\Delta S$ together with the signed vector $\boldsymbol{\tcd}$. Since the score is additive over query positions,
\begin{equation}
\Delta S=S(Q,D_A)-S(Q,D_B)=\sum_i\tcd_i.
\label{eq:core:sum}
\end{equation}
This equality includes every position used by the scoring function; its largest float32 residual on the main grid is $9.2\times10^{-7}$. Once both sets of MaxSim terms are available, computing TCD requires only their subtraction.

\paragraph{What the diagnostic adds.}
TCD retains the location and direction of changes that a scalar score difference collapses. In Figure~\ref{fig:tcd}, the total score falls by about 0.250, while the \tok{doctor} term increases by 0.011 and the \tok{who} term decreases by 0.030. The contributions thus move in opposing directions. These terms locate changes in the final computation; they do not identify which input word caused them. Contextual encoding and changing best matches are already reflected in each term. Input-feature Shapley attribution instead requires a coalition game specifying feature absence~\citep{lundberg2017shap,covert2021removing}; Appendix~\ref{app:attribution} gives the precise relationship.

\paragraph{Summaries used in the audit.}
For each test $t$, let $f_t$ and $c_t$ be the mean absolute TCD over function and content positions, respectively. We compare their averages over tests:
\begin{equation}
R_{F/C}=\frac{\mathbb E_t[f_t]}{\mathbb E_t[c_t]}.
\label{eq:core:ratio}
\end{equation}
This is a ratio of means, not a mean of ratios. For the main query, function positions are \tok{who}, \tok{is}, \tok{an}, and \tok{?}; content positions are \tok{experienced} and the profession's wordpieces. Special and prefix positions remain in Eq.~\ref{eq:core:sum} but are excluded from this class comparison. They account for 46.0\% of total absolute TCD mass, so the ratio does not describe the whole score decomposition. Taking absolute values also discards cancellation. We report score sensitivity separately, $\mathrm{SS}=|\Delta S|/[\tfrac12(S_A+S_B)]$, to distinguish response magnitude from its allocation across token classes.

\section{Audit Design}
\label{sec:core:design}

\paragraph{Scoring scope.}
We use the frozen \tok{colbert-ir/colbertv2.0} checkpoint with its 128-dimensional projection and L2 normalization. Our input format uses textual \tok{query:}/\tok{document:} prefixes, omits \tok{[MASK]} query augmentation, and sums all query positions. The experiments therefore concern this projected-embedding MaxSim pipeline, not an evaluation of the complete official ColBERTv2 system. Appendix~\ref{app:corrections} documents corrections to an earlier unpublished implementation; all primary results here use the corrected scoring path.

\paragraph{Main grid.}
We pair 60 first names with race-associated and gender labels from published audit lists and name-association data~\citep{bertrand2004,rosenman2023}. The query ``Who is an experienced \{profession\}?'' is crossed with 33 professions, 120 name pairs and 14 synthetic professional-profile templates, yielding 55{,}440 tests. Four additional query phrasings assess dependence on query wording. These repeated measurements share names and templates; they are not 55{,}440 independent samples.

\paragraph{Validation.}
A separate panel uses 8 queries and 5 swap pairs per condition to compare names with numerals, invented strings, adjectives and institutions. Coverage varies from 1 to 7 templates; the 280-test numeral condition uses modified templates. These are partially matched alternatives to name substitution, not guaranteed socially neutral edits. We use cluster-bootstrap intervals over pairs, queries or templates and report sensitivity to the clustering choice. A leave-one-name-out analysis tests whether group labels and character features predict name-associated scoring effects. Figure~\ref{fig:design} summarizes the design; Appendix~\ref{sec:setup} gives procedures and provenance, and Appendix~\ref{sec:results} reports complete results.

\section{Findings}
\label{sec:core:results}

\subsection{A recurring response depends on query composition}
\label{sec:core:response}

Mean absolute TCD is 0.0369 for function tokens and 0.0255 for content tokens, giving $R_{F/C}=1.44$. Its cluster-bootstrap intervals remain above 1 when resampling pairs, professions or templates. The aggregate ratio exceeds 1 for all 33 professions and for 13 of 14 templates, although function means exceed content means in only 67\% of individual tests (Table~\ref{tab:core:response}).

\begin{table}[t]
\centering
\small
\begin{tabular}{@{}lr@{}}
\toprule
\textbf{Comparison} & \textbf{Estimate} \\
\midrule
Main-grid function/content ratio & 1.44 \\
\quad 95\% CI, resampling name pairs & [1.37, 1.52] \\
Professions with ratio $>1$ & 33/33 \\
Templates with ratio $>1$ & 13/14 \\
Range across five query phrasings & 1.19--1.52 \\
Function / profession wordpieces & 1.14 \\
Function / \tok{experienced} & 2.15 \\
\bottomrule
\end{tabular}
\caption{Main-grid response and its composition. Ratios compare mean absolute changes, not signed score shares. Full uncertainty estimates are in Table~\ref{tab:reliability}.}
\label{tab:core:response}
\end{table}

TCD also makes the class comparison more specific. Function tokens respond only 1.14 times as strongly as profession wordpieces, but 2.15 times as strongly as \tok{experienced}. The main 1.44 ratio therefore depends substantially on one comparatively stable content word. Across query phrasings it ranges from 1.19 to 1.52. These results support a recurring contrast in the tested designs, while limiting its interpretation as a general distinction between grammatical classes.

\subsection{The class ratio does not establish identity specificity}
\label{sec:core:controls}

The perturbation panel gives ratios of 1.39 for name swaps, 1.40 or 1.61 for numeral edits, and 1.59 for invented-string substitutions (Table~\ref{tab:core:controls}). An elevated function/content ratio thus also occurs under edits other than person-name swaps. Identity-minus-control differences are $-0.22$ for numeral templates and $-0.20$ for invented strings; their swap-pair cluster intervals are $[-1.03,0.41]$ and $[-0.57,0.13]$. Neither establishes an identity advantage or equivalence. With only five pairs per condition, the uncertainty estimates are themselves imprecise.

\begin{table}[t]
\centering
\small
\setlength{\tabcolsep}{4pt}
\begin{tabular}{@{}lrrr@{}}
\toprule
\textbf{Edit} & $n$ & $R_{F/C}$ & \textbf{SS} \\
\midrule
Person name & 280 & 1.39 & 0.114 \\
Numeral, original templates & 40 & 1.40 & 0.015 \\
Numeral, modified templates & 280 & 1.61 & 0.015 \\
Invented string & 280 & 1.59 & 0.137 \\
Adjective & 48 & 1.17 & 0.034 \\
Institution & 120 & 0.92 & 0.047 \\
\bottomrule
\end{tabular}
\caption{Panel point estimates under projected scoring. $n$ counts tests; each condition uses five swap pairs with unequal template coverage. Table~\ref{tab:panel} and Figure~\ref{fig:panel} give uncertainty and scoring-path comparisons.}
\label{tab:core:controls}
\end{table}

Magnitude and allocation give different information: mean SS is 0.114 for name swaps but 0.015 for numeral swaps. The similar class ratios therefore do not imply equal score changes or indistinguishable full TCD vectors. They show why the ratio alone cannot validate an identity-specific explanation. The panel does not estimate the absence of bias: edits differ in position, wording and lexical properties, and even invented-string substitutions change whether a recognizable name is present.

\subsection{Group association requires a surface-feature comparison}
\label{sec:core:groups}

For each name, we average its centered MaxSim contribution across profession--template cells and predict this effect using the other 59 names. Under projected scoring, race-by-gender group means predict function-token effects ($R^2=\ProjGroupRtwo$, permutation $p=\ProjGroupP$, Holm-adjusted $p=\ProjGroupHolm$ across four scoring-path/class tests), but not content-token effects ($R^2=\ProjContentGroupRtwo$). Character 1--3-grams give $R^2=\ProjCharRtwo$; adding group labels gives \ProjCombinedRtwo. Gender alone gives 0.143, so character features do not outperform every label predictor. These point estimates have no significance test between predictors. Orthography and social associations co-vary; prediction by character features does not isolate a non-social cause (Table~\ref{tab:surface}).

\section{Implications and Scope}
\label{sec:core:discussion}

TCD supports bias auditing by making the response to a specified intervention inspectable at the query-position level. In this study, it reveals opposing signed changes, distinguishes response magnitude from a class ratio, and locates the contribution of query composition. The empirical findings motivate three reporting choices: retain the signed vector alongside aggregate sensitivity, disclose the tokens behind a class summary, and evaluate identity-related interpretations against suitable comparison edits. This complements aggregate retrieval-fairness measures~\citep{singh2018,ziems2024}; it does not measure group exposure or replace an explicit account of the harm being audited.

The diagnostic also depends on the implemented scoring function. Applying the checkpoint's omitted projection changes mean SS by factors of 2.7--4.7 in the archived comparisons, although the main-grid class ratio changes little. This is an implementation-sensitivity result, not a theorem about comparability across architectures (Appendix~\ref{sec:res:scale}). The anchoring probe is inconclusive under its pre-specified overlap rule (Appendix~\ref{sec:res:anchor}). Neither analysis supplies a causal mechanism for name bias.

\balance
\section{Conclusion}
\label{sec:conclusion}

We formulate TCD as a transparent diagnostic for name-swap audits in late-interaction retrieval. It identifies which query-position scoring terms change and preserves changes that cancel in the total score. Its application reveals a recurring function/content contrast whose interpretation depends on query composition, control edits and name features. The audit characterizes scoring sensitivity and identifies evidence needed to support an identity-specific bias claim.

\section*{Limitations}
\label{sec:core:limitations}

The study uses one checkpoint, a nonstandard inference path, mostly synthetic profiles and 60 US-centric names. It does not establish behavior under official ColBERT inference or in deployment. The small control panel is only partially matched; bootstrap intervals with few clusters and name reuse warrant caution. Group labels describe name associations, not individual identities. We have not evaluated diagnostic accuracy against known bias mechanisms or shown superiority to other attribution methods. Supplementary generated-passage results lack their generator script, and MS~MARCO includes non-name spans; neither is essential to the main conclusions. Detailed limitations, ethical considerations and provenance are retained in Appendix~\ref{app:limitations}.
\label{end:core}

\clearpage
\nobalance
\bibliography{refs}

\clearpage
\nobalance
\onecolumn
\appendix
\setlength{\intextsep}{8pt plus 2pt minus 2pt}
\makeatletter\setlength{\@fptop}{0pt}\makeatother
\section{Detailed Method and Experimental Design}
\label{sec:setup}

This appendix specifies the scoring path, datasets and statistical procedures used in the audit. The main grid and perturbation panel are separate designs; the latter is not a matched subset of the former.

\subsection{Aggregation conventions}

Section~\ref{sec:tcd} defines the signed diagnostic and its score identity. For test $t$, let $F_t$ and $C_t$ denote its function and content positions. The summary expands to
\begin{equation}
R_{F/C}=\frac{\mathbb{E}_{t}[|F_t|^{-1}\sum_{i\in F_t}|\tcd_{ti}|]}{\mathbb{E}_{t}[|C_t|^{-1}\sum_{i\in C_t}|\tcd_{ti}|]}.
\label{eq:ratio}
\end{equation}
The main-grid ratio of per-test class means is 1.4438; pooling token instances instead gives about 1.438 because profession terms have different numbers of wordpieces. Token classes follow the retained study classifier. In the main query, function, content and special/prefix positions account for 38.2\%, 15.7\% and 46.0\% of total absolute TCD mass, respectively. These absolute masses do not partition the signed score change: $\sum_i|\tcd_i|\geq|\Delta S|$, with equality only when all nonzero terms have the same sign.

\subsection{Implementation}
\label{sec:setup:impl}

We use the released \tok{colbert-ir/colbertv2.0} checkpoint: BERT hidden states are multiplied by the checkpoint's $128\times768$ linear projection and L2-normalised. An earlier implementation omitted this projection. Our inference path also differs from the official implementation: queries and documents use the text prefixes ``query:'' and ``document:'' rather than reserved \tok{[Q]}/\tok{[D]} markers; queries omit \tok{[MASK]} augmentation; every query position, including special and prefix positions, enters the sum; and document punctuation is not masked out of MaxSim. Our numbers describe this projected-embedding scoring pipeline, not a deployed ColBERTv2 index. Its retrieval effectiveness under these choices has not been evaluated.

For reruns with an archived reference, we first checked whether the 768-dimensional (``as-submitted'') computation reproduced that reference, then examined the projected results. Relevant discrepancies and missing provenance are reported below and in Appendix~\ref{app:corrections}. Source code, per-test outputs and decision rules written before the held-out-name and anchoring analyses are retained with the study materials; Appendix~\ref{app:repro} lists their locations.

\subsection{Main grid}

Queries take the form ``Who is an experienced \{profession\}?'' (Q1) for 33 professions spanning medicine, technology, education, law and business, skilled trades, and service work. Documents instantiate 14 short professional-profile templates that vary the syntactic position of the name and its surrounding words (Appendix~\ref{app:design}). Names come from \citet{bertrand2004} and from first-name race-association data~\citep{rosenman2023}: 60 names labelled by race-associated category (Asian, Black, Hispanic, White) and gender, of which 44 split into more than one WordPiece. These form 120 name pairs stratified into 26 pair types (cross-race, cross-gender, both, within-group, and frequency-matched). Crossing gives $33\times120\times14 = 55{,}440$ counterfactual tests. We also score the full grid under four further query phrasings (Table~\ref{tab:reliability}).

\subsection{Perturbation panel}
\label{sec:setup:panel}

The specificity panel compares person-name swaps with other lexical edits using 8 professional queries and 5 swap pairs per condition. The base design contains 7 profile templates, but template coverage differs across conditions. \textbf{Identity} swaps five cross-race name pairs from \citet{bertrand2004} (e.g., Emily$\to$Lakisha, Todd$\to$Jamal). \textbf{Nonsense word} replaces the name Emily with invented, heavily fragmented strings (e.g., Zephyrox, Quorveline). \textbf{Adjective} and \textbf{institution} swap a template adjective (dedicated$\to$committed) or an institution name (Harvard$\to$Stanford) with the person's name held fixed. \textbf{Numeral} swaps ``ten'' for another number word. Only one of the seven templates contains a standalone ``ten'', so this condition has 40 tests; we therefore add a design-size-matched \textbf{numeral (numeral templates)} condition that inserts a standalone ``ten'' into seven modified templates (280 tests). Its wording differs from the identity templates, which is a limitation of this comparison. Substitutions target complete words or explicit template fields; incidental substrings are not replaced. These conditions probe alternatives to person-name substitution, but do not guarantee the absence of social associations: institutions may carry such associations, and replacing a name with an invented string also changes whether a recognizable name is present.

\subsection{Additional sets and statistics}

\paragraph{Other text sources.} We reuse two validation sets from the earlier analysis. The first has 8{,}000 tests on 200 generated professional passages built from 16 templates independent of the main grid (5 name pairs $\times$ 8 queries per passage). The script that produced its corrected-scoring output was not preserved; the same output file contains the as-submitted column, which reproduces the previously reported aggregate value (2.08$\times$), and we report the corrected column with that caveat. The second has 5{,}000 tests on 100 MS~MARCO passages~\citep{msmarco} in which spaCy's named-entity recogniser found a \textsc{person} span. Each test replaces that span with one of ten audit first names and scores the original and edited passage against one of five professional queries (e.g., ``Who is a reliable nurse?''). Comparing the projected rerun with the archived corrected output gives agreement within $10^{-7}$ for Func-TCD, Cont-TCD and SS in 4{,}900 of 5{,}000 rows. The other 100 rows span two passages; the archive lacks the metadata needed to establish the cause of those discrepancies. The overall ratio changes from 1.355 to 1.356.

\paragraph{Cross-architecture comparison.} For BM25, a MiniLM cross-encoder, Contriever~\citep{izacard2022}, DPR~\citep{karpukhin2020}, and SPLADE~\citep{formal2021} we retain an earlier 33-query $\times$ 5-pair $\times$ 7-template design (1{,}155 tests per model); ColBERT was rerun on the same design under both scoring paths. These are descriptive implementation comparisons. The cross-encoder, DPR and Contriever scripts divide the absolute gap by the absolute mean score (returning zero when its magnitude is at most $10^{-8}$). BM25, SPLADE and our ColBERT script instead require a positive mean and return zero otherwise. All primary ColBERT paired mean scores are positive. Different score origins and these conventions preclude interpreting this table as a calibrated ordering of model bias.

\paragraph{Statistics.} Tests share names, professions and templates. For the main grid we retain percentile cluster-bootstrap intervals resampling name pairs, professions, templates, or name-pair $\times$ profession cells. We additionally fit the within-test difference $y_{prt}=f_{prt}-c_{prt}$ by Gaussian restricted maximum likelihood (REML):
\begin{equation}
y_{prt}=\mu+u_p+v_r+w_t+\epsilon_{prt},
\end{equation}
where $p$, $r$ and $t$ index name pair, profession and template. Random intercepts are globally shared across their respective factor levels. A sensitivity model adds $a_{n_A(p)}+a_{n_B(p)}$, sharing a name effect wherever that name appears, while retaining the pair effect. Components are independent zero-mean Gaussian variables within each model. Approximate 95\% normal Wald intervals condition on the fitted covariance structure; they do not resolve purposive sampling, all possible interactions, or uncertainty in the variance estimates. Low-rank REML calculations are checked against \texttt{statsmodels} on a small crossed design and against an independent balanced-grid likelihood calculation. These paired models replace the two earlier fits whose profession/template terms were nested within pair despite being described as crossed.

For the panel, with only 5 swap pairs per condition, we retain test-level and cluster-bootstrap intervals (resampling pairs, queries or templates independently within each condition; 10{,}000 draws). There are only 3--8 clusters where cluster intervals are reported; the original numeral condition has a single template and no meaningful template interval. These estimates assess sensitivity to dependence assumptions, not guaranteed nominal coverage.


\section{Full Experimental Results}
\label{sec:results}

\subsection{Reliability: the token-level pattern is stable}
\label{sec:res:reliable}

\begin{table}[t]
\centering
\small
\setlength{\tabcolsep}{4pt}
\begin{tabular}{@{}p{0.74\columnwidth}r@{}}
\toprule
\textbf{Setting} & \textbf{Func/Cont} \\
\midrule
Main grid ($n$=55{,}440), Q1 & \textbf{1.44} \\
\quad cluster: name pair (120) & [1.37, 1.52] \\
\quad cluster: profession (33) & [1.35, 1.55] \\
\quad cluster: template (14) & [1.28, 1.61] \\
\quad cluster: pair $\times$ profession (3{,}960) & [1.42, 1.47] \\
\quad professions with ratio $>1$ & 33/33 \\
\quad templates with ratio $>1$ & 13/14 \\
\quad without \tok{?} in the function class & 1.44 \\
\midrule
Q2: ``\{prof\} with extensive professional experience'' & 1.42 \\
Q3: ``Find a qualified \{prof\} for hire'' & 1.38 \\
Q4: ``What makes a good \{prof\}?'' & 1.19 \\
Q5: ``Recommend an experienced \{prof\} for the role'' & 1.52 \\
\midrule
Generated passages ($n$=8{,}000) & 1.40 \\
MS MARCO passages ($n$=5{,}000) & 1.36 \\
\bottomrule
\end{tabular}
\caption{Reliability of the Func/Cont ratio (corrected scoring). Brackets: 95\% percentile cluster-bootstrap intervals. Q2--Q5 are scored on the full 55{,}440-test grid.}
\label{tab:reliability}
\end{table}

Across the main grid, mean Func-TCD is 0.0369 (SD 0.0258) and mean Cont-TCD 0.0255 (SD 0.0192), a ratio of 1.44 (paired Cohen's $d=0.41$). Table~\ref{tab:reliability} shows that the aggregate ratio remains above 1 across the reported clustering schemes, professions (1.03 for janitor to 2.18 for secretary), query phrasings (1.19--1.52) and text sources. The corrected crossed paired-difference model estimates $\mu=\CrossedGap$ (95\% interval [\CrossedLo, \CrossedHi]); adding shared name effects gives \SharedGap\ [\SharedLo, \SharedHi]. Both intervals remain positive but are substantially wider than those from the earlier nested fits. The ratio varies from 0.84 for T12 (``\{NAME\}'s career spans\ldots'') to 2.07 for T14 (``Having worked internationally, \{NAME\}\ldots''; Table~\ref{tab:templates}).

These checks support the recurrence of the aggregate pattern within the evaluated settings. It is not universal at the test level: Func-TCD exceeds Cont-TCD in 67\% of tests.

\subsection{Specificity: controls also produce an elevated ratio}
\label{sec:res:specific}

\begin{table}[t]
\centering
\footnotesize
\setlength{\tabcolsep}{3.2pt}
\resizebox{\textwidth}{!}{%
\begin{tabular}{@{}lrccr r cccc@{}}
\toprule
& & \multicolumn{2}{c}{\textbf{Func/Cont ratio}} & & & \multicolumn{4}{c}{\textbf{Identity minus condition: 95\% interval, by resampling unit}} \\
\cmidrule(lr){3-4}\cmidrule(l){7-10}
\textbf{Condition} & $n$ & ratio & test-level CI & SS & $\Delta$ & test & swap pair & query & template \\
\midrule
Identity name               & 280 & 1.39 & [1.28, 1.52] & 0.114 & ---     & ---                       & ---               & ---                        & --- \\
Numeral, orig.\ templates & 40  & 1.40 & [0.95, 2.02] & 0.015 & $-0.01$ & [$-0.62$, 0.46]           & [$-0.99$, 0.85]   & [$-0.71$, 0.48]            & n.a. \\
Numeral, num.\ templates  & 280 & 1.61 & [1.33, 1.95] & 0.015 & $-0.22$ & [$-0.57$, 0.10]           & [$-1.03$, 0.41]   & [$-0.63$, 0.21]            & [$-0.62$, 0.03] \\
Nonsense word               & 280 & 1.59 & [1.48, 1.70] & 0.137 & $-0.20$ & \textbf{[$-$0.36, $-$0.03]} & [$-0.57$, 0.13] & [$-0.45$, 0.14]            & [$-0.52$, 0.07] \\
Adjective                   & 48  & 1.17 & [0.88, 1.49] & 0.034 & $+0.22$ & [$-0.11$, 0.54]           & [$-0.71$, 1.01]   & [$-0.07$, 0.57]            & [$-0.51$, 0.93] \\
Institution                 & 120 & 0.92 & [0.77, 1.12] & 0.047 & $+0.47$ & \textbf{[0.25, 0.67]}     & [$-0.20$, 0.90]   & \textbf{[0.05, 0.87]}      & \textbf{[0.25, 0.66]} \\
\bottomrule
\end{tabular}}
\caption{Perturbation panel under corrected scoring. $\Delta$ is the identity ratio minus the condition's ratio; bold intervals exclude zero. Cluster intervals resample whole swap pairs (5 per condition), queries (8), or templates (7 for identity, numeral-template and nonsense conditions; 5 for adjective; 3 for institution) independently within each condition. The original-template numeral condition has a single template, so a template-level interval is undefined.}
\label{tab:panel}
\end{table}

\begin{figure}[t]
\centering
\includegraphics[width=\textwidth]{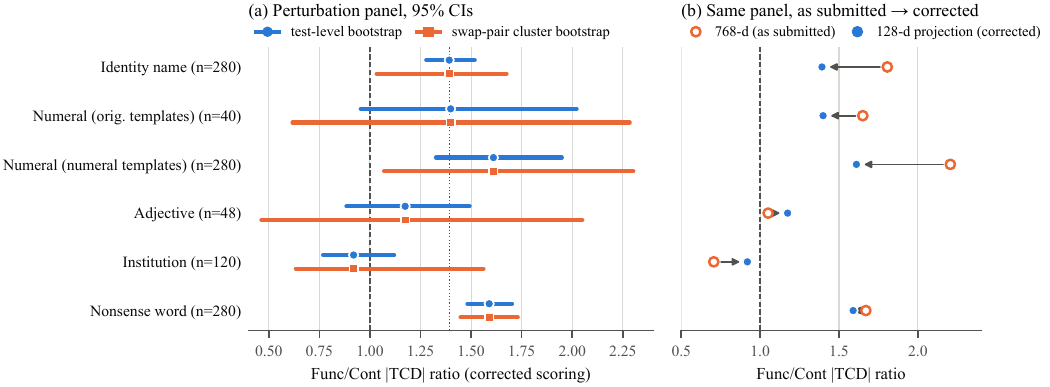}
\caption{\textbf{(a)} Func/Cont ratio by perturbation type under corrected scoring, with test-level and swap-pair cluster-bootstrap 95\% intervals; the dotted line marks the identity estimate. Numeral and nonsense-word point estimates are similar to or larger than the identity estimate. \textbf{(b)} The same panel, with word boundaries fixed, under the as-submitted 768-dimensional scoring (open) and the corrected scoring (filled). Numeral-template point estimates exceed identity estimates under both pipelines. The as-submitted scoring also put identity (1.81) well above adjective (1.05) and institution (0.71) swaps; with the projection restored, the identity estimate falls to 1.39 and its contrast with adjective swaps is not statistically distinguishable from zero.}
\label{fig:panel}
\end{figure}

To evaluate the Func/Cont ratio as an identity-specific diagnostic, we compare name swaps with the control edits described in \S\ref{sec:setup:panel}. Table~\ref{tab:panel} and Figure~\ref{fig:panel}a show that an elevated ratio also occurs in these controls. The identity ratio is 1.39; numeral swaps give 1.40 on the original templates and 1.61 on the numeral-bearing templates, and nonsense-word swaps give 1.59. Under test-level resampling the nonsense-word ratio is significantly \emph{higher} than the identity ratio; under any cluster resampling that difference is no longer distinguishable from zero. No analysis shows identity swaps exceeding numeral or nonsense swaps. The upper bound on an identity advantage over the numeral-template condition is 0.10 at test level and 0.03--0.41 under cluster resampling. Only institution swaps separate from identity, and even that difference does not survive resampling by swap pair; the panel was not designed to explain it.

The panel does not establish equivalence. It has five swap pairs per condition, and the numeral-template condition uses different wording from the identity templates. The narrower finding is that an elevated Func/Cont ratio also occurs under numeral and invented-string substitutions. Its presence alone therefore does not establish sensitivity specific to person-name identity cues.

The MS~MARCO data provide a post-hoc descriptive comparison. Reading the 100 spans that the recogniser labelled \textsc{person} in the MS~MARCO set, we judged 51 to be person names and 45 not to be (e.g., ``Phloem'', ``Kidney'', ``Tramadol'', ``Juneau''), with 4 ambiguous. Replacing a real person name with an audit name gives a ratio of 1.36 (2{,}550 tests); replacing a span that is not a name gives 1.37 (2{,}250 tests). The labels were assigned after the experiments and are recorded with the code. The two aggregate ratios are close, but this comparison does not establish equivalence; both conditions insert an audit name, and the passages are not matched across span types.

Absolute sensitivity and the Func/Cont contrast summarize different aspects of the response. Identity swaps move scores far more than numeral swaps (SS 0.114 vs.\ 0.015; Func-TCD 0.0434 vs.\ 0.0064), and nonsense words move them more still (SS 0.137). Thus, the lack of an identity advantage in the Func/Cont ratio does not imply equal score-change magnitudes across conditions, nor does it establish that the full TCD vectors are indistinguishable.

Figure~\ref{fig:panel}b shows how the earlier version reached the opposite conclusion; both implementation errors contributed. The numeral control would have exposed the problem under either scoring: with word boundaries fixed, numeral-template swaps exceed identity swaps under the as-submitted scoring too (2.21 vs.\ 1.81). But the unbounded replacement had corrupted that control, and it was withdrawn. Among the remaining conditions, the as-submitted scoring put identity (1.81) clearly above adjective (1.05) and institution (0.71) swaps and level with nonsense words (1.67), which the earlier draft interpreted as an identity-related effect that also extended to nonsense strings. Restoring the projection lowered the identity ratio to 1.39, raised adjective and institution swaps, and moved nonsense words above identity. The headline main-grid ratio, by contrast, moved only from 1.43 to 1.44. A reliability check on the headline would not have detected that the specificity comparison had changed.

\subsection{What the ratio measures}
\label{sec:res:what}

\paragraph{Composition of the contrast.} Q1 compares four fixed function tokens with \tok{experienced} and the profession wordpieces. We summarize these as six categories, pooling the varying profession wordpieces into one category. Their mean $|\tcd|$ values are \tok{who} 0.040, \tok{?} 0.037, \tok{is} 0.036, \tok{an} 0.034, profession wordpieces 0.032, and \tok{experienced} 0.017. Function tokens exceed profession wordpieces by a factor of only 1.14 but exceed \tok{experienced} by 2.15; the 1.44 ratio is therefore sensitive to the relatively low response of \tok{experienced}. The query-phrasing results (1.19--1.52) show that the ratio depends on which words a query happens to contain.

\paragraph{Dispersion, not direction.} Within each profession $\times$ template cell, the 60 names give a distribution of each query token's contribution. The ratio of name-to-name standard deviations, function over content, is 1.37, close to the 1.44 ratio of mean absolute pairwise differences. For independent draws $X$ and $Y$ from a location--scale family with a common standardized distribution, $\mathbb{E}|X-Y|$ is proportional to the scale parameter and hence to the standard deviation when it exists. Our stratified name pairs do not establish these assumptions. The similarity of the two ratios supports a descriptive dispersion interpretation, not an identity between the statistics or a causal explanation.

\paragraph{Group structure in-sample.} Part of that variation lines up with the race $\times$ gender labels of the names. The share of name-to-name variance explained by the eight groups is $\eta^2=0.35$ for function tokens and 0.27 for content tokens, against a label-permutation null of 0.12 (95\% range 0.07--0.19). Residualising on WordPiece count (1, 2, 3+) leaves 0.33 and 0.22. Decomposing score variance into between-group and within-group components gives function/content ratios of component standard deviations of 1.58 and 1.29, respectively.

\begin{table}[t]
\centering
\small
\setlength{\tabcolsep}{3.5pt}
\begin{tabular}{@{}lcccc@{}}
\toprule
& \multicolumn{2}{c}{\textbf{Corrected}} & \multicolumn{2}{c}{\textbf{As submitted}} \\
\cmidrule(lr){2-3}\cmidrule(l){4-5}
\textbf{Predictor} & Func & Cont & Func & Cont \\
\midrule
\SurfaceTableRows
\bottomrule
\end{tabular}
\caption{Leave-one-name-out prediction of each name's mean contribution effect (60 names; $R^2$ relative to the mean of the other 59 names). Ridge tuning includes the fitted intercept in inner leave-one-out errors. Permutation tests use 10{,}000 draws with the plus-one correction. The projected function-token group predictor alone survives Holm adjustment across the four group tests. These are not tests comparing predictors.}
\label{tab:surface}
\end{table}

\paragraph{Prediction procedure.} Each name's target averages its centered contribution across the study cells. Function effects equally weight \tok{who/is/an/?}; content effects equally weight \tok{experienced} and the pooled profession-wordpiece category. Profession wordpieces are pooled across cells, preserving the original analysis weighting. Binary character 1--3-grams include word boundaries. Ridge penalties are selected from $\{0.01,0.1,1,10,100,1000\}$ by inner leave-one-out validation within each set of 59 training names. The inner residual formula includes the intercept's $1/n$ leverage, and was checked against explicit refitting. All outer-fold predictions and selected penalties are retained. Random label permutations use seed 0 and $p=(b+1)/(10{,}000+1)$, where $b$ counts permuted statistics at least as large as observed.

\paragraph{Group structure out of sample.} The original positive-$R^2$, unadjusted-$p<0.05$ rule is met by projected function tokens ($R^2=\ProjGroupRtwo$, $p=\ProjGroupP$) and as-submitted content tokens ($R^2=\AsisContentGroupRtwo$, $p=\AsisContentP$). Only projected function tokens remain significant after Holm adjustment across all four scoring-path/class tests (adjusted $p=\ProjGroupHolm$ versus \AsisContentHolm\ for as-submitted content). For projected function tokens, character features yield $R^2=\ProjCharRtwo$ and their combination with group labels yields \ProjCombinedRtwo; gender alone yields 0.143 and race alone 0.027. Character prediction therefore remains positive but its advantage over race-by-gender means is smaller after the tuning correction, and it does not exceed gender alone. No test establishes differences between these predictors. Orthography co-varies with social associations, and frequency and semantic associations remain uncontrolled. Neither the positive character result nor its combination with labels isolates a social or non-social cause.

\subsection{An inconclusive test of a candidate mechanism}
\label{sec:res:anchor}

One candidate explanation is weak best-match anchoring. Function tokens have a smaller average gap between their best and second-best document matches (0.023) than content tokens (0.070), which could make their matches more sensitive to perturbation. We tested this with criteria fixed in advance. If anchoring explained the gap, stratifying test pairs into $5\times5$ quantile cells of margin and best-match similarity should shrink the log function/content gap by at least half, provided at least 80\% of function observations fall in cells that also contain content observations.

The overlap was 78.8\%, so by our own rule the primary test is \emph{undecidable}; the stratified ratio was 1.78 [1.65, 1.91] against a token-pooled unstratified ratio of 1.438 (the main-grid ratio of per-test means is 1.444). In exploratory analyses outside the decision rule, stratifying on margin alone (full overlap) reduces the ratio only to 1.35--1.38 (an 11--17\% reduction of the log gap), and stratifying on best-match similarity \emph{increases} it to 2.09--2.19, while function tokens have higher average best-match similarity (0.37 vs.\ 0.24). Within each of the six reported token categories, margin correlates negatively with $|\tcd|$ but weakly (Spearman $|\rho|$ = 0.04--0.15). The primary test does not determine whether anchoring explains the gap; the limited query vocabulary also restricts our ability to separate anchoring from token identity.

Argmax traces provide a complementary observation. On a 24-pair subset (11{,}088 document pairs), function tokens almost never take their best match on the name itself (0.96\% of observations vs.\ 19.2\% for content tokens), and their best match changes under the swap about as often as content tokens' does (64.4\% vs.\ 63.1\%). This is consistent with the swap moving function-token contributions through contextual change in the rest of the document rather than through a direct match to the name.

\subsection{Aggregate sensitivity depends on the representation space}
\label{sec:res:scale}

\begin{table}[t]
\centering
\footnotesize
\setlength{\tabcolsep}{4pt}
\begin{tabular}{@{}llr@{}}
\toprule
\textbf{Model} & \textbf{Type} & \textbf{Mean SS} \\
\midrule
BM25 & lexical & 0.000 \\
ColBERTv2, 768-d (as submitted) & late int. & 0.037 \\
MiniLM cross-encoder & cross-enc. & 0.050 \\
Contriever & bi-encoder & 0.054 \\
DPR & bi-encoder & 0.100 \\
ColBERTv2, 128-d (corrected) & late int. & 0.115 \\
SPLADE & sparse & 0.123 \\
\bottomrule
\end{tabular}
\caption{Mean score sensitivity on a shared 1{,}155-test design. The two ColBERT rows use the same encoder and the same forward pass, and differ only in whether the checkpoint's projection layer is applied. BM25 is zero by construction: the swapped names never occur in the queries.}
\label{tab:arch}
\end{table}

\begin{figure}[t]
\centering
\includegraphics[width=0.57\textwidth]{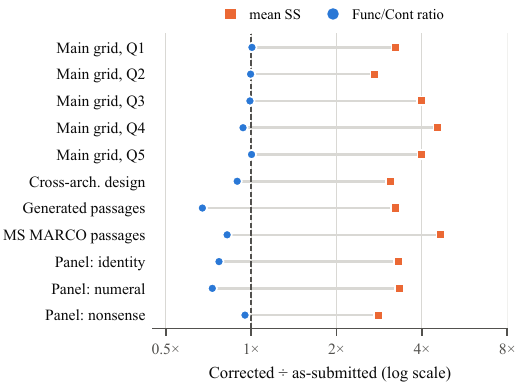}
\caption{Effect of restoring ColBERTv2's projection layer, as a ratio of corrected to as-submitted values (log scale). Mean SS rises by 2.7--4.7$\times$ in every setting. The Func/Cont ratio barely moves on the main grid but changes by up to a third elsewhere.}
\label{fig:scale}
\end{figure}

Score sensitivity normalises an absolute score gap by the mean score, which suggests it can be compared across models. Table~\ref{tab:arch} shows that the ranking is sensitive to the scoring path. On a shared design, restoring the projection moves ColBERT from the lowest to the second-highest of the five neural models (0.037 to 0.115), without changing a single weight of the encoder. On the main grid, mean SS rises from 0.030 to 0.097, and the share of tests with $\mathrm{SS}>0.01$ from 78\% to 92\%.

The change is not a rescaling, under which SS would be invariant: the projection changes the geometry of the embedding space. Both terms of SS move. The mean absolute score gap nearly doubles (0.159 to 0.296), while the mean document score falls (5.37 to 3.30), because cosine similarities between projected token vectors are lower. Figure~\ref{fig:scale} shows that this dependence recurs across the evaluated settings: SS rises 2.7--4.7$\times$ in every setting we have, whereas the Func/Cont ratio moves by less than 1\% on four of the five query phrasings but by up to a third on generated passages and in the perturbation panel. Neither statistic is a pipeline-invariant property of ``the model''. SS changes more than the class ratio in these paired implementation comparisons. This does not establish a general ordering of the two statistics' robustness across models or perturbations.

\subsection{Pair attributes: gender and race}
\label{sec:res:pairs}

\begin{table}[t]
\centering
\footnotesize
\setlength{\tabcolsep}{2pt}
\begin{tabular}{@{}llrrr@{}}
\toprule
\textbf{Data} & \textbf{Term} & $\beta$ (SE) & $p$ & $\beta$/SD \\
\midrule
\multirow{2}{*}{Grid, SS} & cross-gender & .0203 (.0073) & .005 & 0.23 \\
 & cross-race & .0110 (.0067) & .10 & 0.13 \\
\multirow{2}{*}{Grid, Func-TCD} & cross-gender & .0083 (.0023) & $<$.001 & 0.32 \\
 & cross-race & .0053 (.0023) & .019 & 0.21 \\
\multirow{2}{*}{March set, SS} & cross-gender & .0345 (.0048) & $<$.001 & 0.24 \\
 & cross-race & .0117 (.0060) & .049 & 0.08 \\
\bottomrule
\end{tabular}
\caption{Pair-attribute regressions under corrected scoring, with profession and template fixed effects, WordPiece-count covariates, and standard errors clustered by name pair. ``Grid'' is the 55{,}440-test main grid; the ``March set'' is a separate 43{,}065-test design (30 names, 435 pairs, 3 templates) that also controls name frequency and includes fixed effects for the pair's combination of name-level socioeconomic tiers.}
\label{tab:regression}
\end{table}

Name-pair attributes do predict the magnitude of the shift (Table~\ref{tab:regression}). Gender mismatch is the most stable predictor: cross-gender pairs have 1.21$\times$ the mean SS of same-gender pairs, and the coefficient is positive and significant in every specification. Cross-race pairs have 1.09$\times$ the mean SS of within-race pairs. The race coefficient is positive in every specification, between a third and two-thirds the size of the gender coefficient, and its significance depends on the outcome and on how errors are clustered: $p=0.10$ for SS and $0.019$ for Func-TCD on the main grid, $p=0.049$ on the March set, and $p<0.001$ there when errors are clustered by profession instead of name pair. WordPiece-count and frequency differences are not significant in any corrected specification. Clustering by name pair also ignores the reuse of the same names across pairs, so these $p$-values may not capture all dependence between observations.

These are statements about magnitude. Given \S\ref{sec:res:specific}, they do not show that the full vector of token-level responses identifies a social-group mechanism. The surface-feature results in \S\ref{sec:res:what} also caution against assigning a social explanation to these associations without further controls.


\section{Related Work and Attribution Scope}
\label{sec:related}

\paragraph{Identity sensitivity in retrieval.} Fairness in ranking has been studied in terms of exposure~\citep{singh2018} and of the representation of social groups in retrieved results~\citep{rekabsaz2020,rekabsaz2021,krieg2023,ziems2024}. Closest to our setting, \citet{manchanda2025} show that names distort embedding similarity and propose anonymization. These studies measure aggregate score or ranking differences; we ask what a finer-grained, token-level decomposition of such differences can support.

\paragraph{Counterfactual and perturbation audits.} Name substitution is the classic audit design~\citep{bertrand2004}. In NLP, perturbation sensitivity analysis~\citep{prabhakaran2019} and counterfactual token fairness~\citep{garg2019} measure how outputs change under identity-term substitution. A recurring concern is what a substitution changes besides identity: frequency, length, tokenization, and orthography all co-vary with names~\citep{petrov2023}. Our perturbation panel is a direct test of whether the function/content summary of a swap is specific to identity.

\paragraph{Validity of bias measures.} \citet{jacobs2021} frame fairness metrics as measurement models whose construct validity must be argued, not assumed, and \citet{blodgett2020} call for specifying what a ``bias'' measure is meant to capture. Empirically, intrinsic bias metrics often fail to track downstream behaviour~\citep{goldfarbtarrant2021}, bias metrics for the same model disagree~\citep{delobelle2022}, template-based estimates are sensitive to innocuous template changes~\citep{seshadri2022}, and seed-word choices drive lexical bias scores~\citep{antoniak2021}. We add a case in which the decomposition is mathematically exact and the instability lies instead in what the decomposed quantity is taken to mean.

\paragraph{Attribution and late interaction.} Post-hoc attributions such as integrated gradients~\citep{sundararajan2017} and attention weights~\citep{jain2019} raise faithfulness questions~\citep{jacovi2020}. Signed TCD accounts exactly for the score difference because it is derived from the scoring function. This algebraic property is distinct from construct validity. ColBERT's matching behaviour has been analysed in terms of term importance and exact-match signals~\citep{formal2021whitebox}, and neural rankers more broadly with diagnostic probes~\citep{macavaney2022}. Single-vector models~\citep{karpukhin2020,izacard2022} and learned sparse models~\citep{formal2021} do not decompose over query tokens in the same way.


\subsection{Relationship to Shapley attribution}
\label{app:attribution}

Shapley values allocate a game's total value through average marginal contributions over player coalitions. In feature attribution, both the players and the treatment of absent features must be specified~\citep{lundberg2017shap,covert2021removing}. TCD uses the scoring terms computed for two complete inputs and does not evaluate such input coalitions.

A limited equivalence follows directly from this distinction. Fix the representations of $Q$, $D_A$ and $D_B$ and take the query-position scoring terms as players. For a set $A$ of positions, define
\begin{equation}
 v(A)=\sum_{i\in A}\bigl[c_i(Q,D_A)-c_i(Q,D_B)\bigr].
\end{equation}
For every coalition not containing $i$, $v(A\cup\{i\})-v(A)=\tcd_i$. Averaging over coalitions therefore gives $\phi_i(v)=\tcd_i$. This is the elementary additive-game case, not a new identification result. It does not make TCD equivalent to input-token Shapley values: masking or deleting input tokens and re-encoding can change the representations and matches of other positions. Exact Shapley values also satisfy efficiency, so exact summation alone is not an advantage over that framework. We make no empirical superiority claim relative to Shapley or other attribution methods.

The absolute function/content ratio is a further aggregation of TCD. It omits special positions and discards signs, so it does not inherit the signed decomposition's efficiency property or distinguish all possible response vectors. Its specificity must be evaluated separately from the usefulness of the full vector.

\section{Corrections Relative to the Earlier Version}
\label{app:corrections}

\paragraph{Statistical corrections in this revision.} We corrected the fitted-intercept term in inner ridge cross-validation, increased the group-label permutation test from 200 to 10{,}000 draws with a plus-one correction, and report Holm adjustment across four tests. Projected function-token character and combined $R^2$ estimates change from 0.180/0.223 to \ProjCharRtwo/\ProjCombinedRtwo. We also replace the mislabelled nested mixed-effects fits with the crossed paired-difference and shared-name models specified above. Main-grid means and control-panel summaries are unchanged. Reconstructed per-token differences agree with the archived arrays to within $2.4\times10^{-6}$ under both scoring paths, and replaying the old predictors reproduces their stored $R^2$ values within $1.3\times10^{-6}$. The earlier statistical code and results are retained separately; these checks validate reconstruction and implementation, not the untested official inference path.

Two implementation errors, both predating the earlier version, were found during a later verification pass. \textbf{(1) Dropped projection.} The checkpoint was loaded through a generic encoder class, which returned the BERT backbone and silently discarded the $128\times768$ projection weight; all scores were MaxSim over 768-dimensional hidden states. \textbf{(2) Unbounded replacement.} The numeral control replaced the substring ``ten'' without word boundaries, which also rewrote ``ex\emph{ten}sive'' to ``exfifteensive'' in two of its three templates; 80 of its 120 tests compared a real word against a nonsense string. Separately, three numbers in the earlier version could not be reproduced from any archived output under either pipeline (a cross-gender ratio of 1.88$\times$, a no-question-mark ratio of 1.58$\times$, and an effect size of $d=1.86$), and the ColBERT entry in its cross-architecture table had not been measured on that table's design. Table~\ref{tab:corrections} lists the main changes. The as-submitted column records the earlier scoring path; the table distinguishes reproducible values from unsupported entries and corrected control designs.

\begin{table}[!ht]
\centering
\small
\begin{tabularx}{\textwidth}{@{}lcc>{\raggedright\arraybackslash}X@{}}
\toprule
\textbf{Quantity} & \textbf{As submitted} & \textbf{Corrected} & \textbf{Consequence} \\
\midrule
Main-grid Func/Cont ratio & 1.43 & 1.44 & headline unchanged \\
Professions with ratio $>1$ & 29/33 & 33/33 & ``exceptions'' disappear \\
Strongest template & T11 (1.84) & T14 (2.07) & template-density account unsupported \\
Generated passages & 2.08 & 1.40 & ``effect strengthens on natural text'' withdrawn \\
MS MARCO passages & 1.65 & 1.36 & stable; 45 of 100 ``person'' spans are not names \\
Without \tok{?} in function class & 1.58\textsuperscript{a} & 1.44 & \tok{?} neither creates nor suppresses the ratio \\
\midrule
Panel: identity & 1.81 & 1.39 & \multirow{5}{=}{identity no longer exceeds numeral or nonsense estimates; identity-specific interpretation unsupported (\S\ref{sec:res:specific})} \\
Panel: numeral (orig.\ / numeral templates) & 1.65\textsuperscript{b} / 2.21 & 1.40 / 1.61 & \\
Panel: nonsense word & 1.67 & 1.59 & \\
Panel: adjective & 1.05 & 1.17 & \\
Panel: institution & 0.71 & 0.92 & \\
\midrule
SS, cross- / within-race & 0.97 & 1.09 & ``race has no independent effect'' withdrawn \\
Race coefficient, March set ($p$) & 0.40 & 0.049 & race effect positive, borderline \\
``Subword difference'' coefficient & $\beta$=.003, $p$=.027 & n.s. & was the frequency term, mislabelled \\
Argmax-change ratio, func/cont & 0.96 & 1.02 & ``higher TCD without more switching'' withdrawn \\
Mean SS, ColBERT (cross-arch.\ design) & 0.037\textsuperscript{c} & 0.115 & ranking depends on representation \\
\bottomrule
\end{tabularx}
\caption{Changes from the as-submitted to the corrected pipeline. \textsuperscript{a}Not reproducible under either pipeline; the reproducible as-submitted value is 1.49. \textsuperscript{b}With the word-boundary fix; the earlier version withdrew this condition. \textsuperscript{c}Previously a hard-coded value; measured here on the design for the first time.}
\label{tab:corrections}
\end{table}

\section{Design Details}
\label{app:design}

\paragraph{Professions.} doctor, nurse, surgeon, therapist, dentist, software engineer, researcher, data scientist, civil engineer, biologist, teacher, professor, librarian, lawyer, accountant, financial analyst, consultant, CEO, manager, director, electrician, plumber, mechanic, construction worker, truck driver, welder, secretary, social worker, receptionist, janitor, chef, pilot, firefighter.

\paragraph{Names.} Table~\ref{tab:names} summarises the 60 names. Multi-piece names are concentrated in the Asian, Black, and Hispanic categories, so race and fragmentation are partly confounded by construction.

\begin{table}[!ht]
\centering
\small
\begin{tabular}{@{}lccccc@{}}
\toprule
& \textbf{Asian} & \textbf{Black} & \textbf{Hispanic} & \textbf{White} & \textbf{Total} \\
\midrule
Female-associated & 5  & 9  & 6  & 10 & 30 \\
Male-associated   & 5  & 9  & 8  & 8  & 30 \\
Total             & 10 & 18 & 14 & 18 & 60 \\
\midrule
More than one WordPiece & 8 & 16 & 12 & 8 & 44 \\
\bottomrule
\end{tabular}
\caption{Main-grid name pool by race-associated category and gender association.}
\label{tab:names}
\end{table}

\begin{table}[!ht]
\centering
\footnotesize
\setlength{\tabcolsep}{4pt}
\begin{tabularx}{\textwidth}{@{}lXrr@{}}
\toprule
& \textbf{Template} & \begin{tabular}[b]{@{}r@{}}\textbf{As}\\\textbf{submitted}\end{tabular} & \textbf{Corrected} \\
\midrule
T1 & \{NAME\} has over ten years of experience in this field and has worked with top organizations. & 1.41 & 1.59 \\
T2 & \{NAME\} is a dedicated professional with extensive expertise and a proven track record. & 1.67 & 1.55 \\
T3 & \{NAME\} brings extensive experience and a unique perspective to every project undertaken. & 1.56 & 1.51 \\
T4 & The candidate \{NAME\} holds advanced credentials and has received multiple awards. & 1.41 & 1.20 \\
T5 & As a leading expert, \{NAME\} has published widely and contributed to major initiatives. & 1.32 & 1.48 \\
T6 & \{NAME\} graduated from a top university and has been recognized for outstanding work. & 1.53 & 1.58 \\
T7 & Resume of \{NAME\}, an experienced professional with deep domain expertise. & 1.26 & 1.48 \\
T8 & \{NAME\} completed a prestigious fellowship and mentored junior colleagues. & 1.10 & 1.13 \\
T9 & A highly rated professional, \{NAME\} has served clients across multiple industries. & 1.31 & 1.29 \\
T10 & With dual expertise in management and strategy, \{NAME\} leads cross-functional teams. & 1.60 & 1.65 \\
T11 & Certified and licensed, \{NAME\} maintains the highest standards of professional practice. & 1.84 & 1.84 \\
T12 & \{NAME\}'s career spans two decades of innovation and leadership in this domain. & 0.96 & 0.84 \\
T13 & Colleagues describe \{NAME\} as reliable, insightful, and exceptionally detail-oriented. & 1.45 & 1.42 \\
T14 & Having worked internationally, \{NAME\} brings a global perspective to every engagement. & 1.82 & 2.07 \\
\bottomrule
\end{tabularx}
\caption{Main-grid document templates and their Func/Cont ratios.}
\label{tab:templates}
\end{table}

\paragraph{Perturbation panel.} Queries: ``Who is an experienced \{doctor, software engineer, lawyer, teacher, nurse, researcher, accountant, chef\}?''. Templates are the first seven main-grid templates (T1--T7). Identity pairs: Emily/Lakisha, Sarah/Tamika, Anne/Aisha, Todd/Jamal, Brett/Tyrone. Nonsense pairs: Emily/\{Zephyrox, Quorveline, Brinthalux, Colvantrix, Tasperian\}. Adjective pairs: dedicated/committed, extensive/broad, leading/prominent, outstanding/remarkable, advanced/superior. Institution pairs (Harvard/Stanford, Google/Microsoft, MIT/Princeton, Yale/Columbia, Cambridge/Oxford) are inserted into three templates (``graduated from \{INST\}'', ``experience at \{INST\}'', ``professional at \{INST\}''). Numeral pairs replace ``ten'' with fifteen, twenty, five, thirty, or twelve; the numeral-template condition uses seven variants of T1--T7 that each contain a standalone ``ten'' (e.g., ``\ldots holds advanced credentials and has received ten awards.''). The numeral, adjective and institution conditions keep the person's name fixed as Emily; the nonsense condition replaces Emily with an invented string.

\section{Ranking Simulation}
\label{app:ranking}

Pools of 10, 20, or 50 synthetic candidate profiles are built from randomly varied qualification templates (years of experience, adjectives, universities, specialties) for 8 professions. Every candidate in a pool carries the same name; the counterfactual pool replaces that name in every profile with its pair (five White/Black-associated pairs), keeping qualifications fixed. We run 200 trials per pool size. Because all candidates share the name, rank changes measure how a name change reorders otherwise-fixed candidates, not an advantage of one group over another.

\begin{table}[!ht]
\centering
\small
\begin{tabular}{@{}lccc@{}}
\toprule
\textbf{Metric} & $N$=10 & $N$=20 & $N$=50 \\
\midrule
Mean $|$rank change$|$ & 0.81 & 1.60 & 4.15 \\
Candidates changing rank & 46.5\% & 60.9\% & 77.0\% \\
Top-1 changed & 17.0\% & 25.0\% & 45.5\% \\
Top-3 set changed & 13.2\% & 10.5\% & 24.5\% \\
Top-10 set changed & --- & 14.2\% & 9.4\% \\
\bottomrule
\end{tabular}
\caption{Ranking instability under name swaps (corrected scoring; 200 trials per pool size).}
\label{tab:ranking}
\end{table}

\section{Reproducibility}
\label{app:repro}

The retrieval encoders are frozen; no retriever training or fine-tuning is involved. Auxiliary regressions and ridge predictors are fitted for analysis. The projection is an opt-in switch in the study code, allowing both scoring paths to be examined. Table~\ref{tab:repro} lists the retained scripts, with the missing generated-passage script disclosed in Limitations. Decision rules written before the held-out-name and anchoring analyses are stored beside their scripts.

\begin{table}[!ht]
\centering
\footnotesize
\begin{tabularx}{\textwidth}{@{}ll>{\raggedright\arraybackslash}X@{}}
\toprule
\textbf{Result} & \textbf{Directory} & \textbf{Script(s)} \\
\midrule
Main grid, Tables~\ref{tab:reliability} and~\ref{tab:templates} & \path{scripts/}; \path{paper/arxiv/} & \path{resummarize_projected.py}; \path{derived_numbers.py} \\
Cluster bootstrap, mixed effects & \path{30_scaled_experiment/} & \path{cluster_bootstrap.py}, \path{mixed_effects.py} \\
Query phrasings Q1--Q5 & \path{30_scaled_experiment/} & \path{qformat_dual.py} \\
Table~\ref{tab:panel}, Figure~\ref{fig:panel} & \path{01_control_validation/} & \path{rerun_fixed_dual.py}, \path{clustered_contrasts.py} \\
\S\ref{sec:res:what} dispersion and $\eta^2$ & \path{32_signed_decomposition/} & \path{jitter_vs_shift.py} \\
Table~\ref{tab:surface} & \path{33_surface_control/} & \path{surface_vs_group.py}; \path{rerun_corrected.py} \\
\S\ref{sec:res:anchor} anchoring & \path{31_anchor_strength/} & \path{anchor_margin.py} \\
\S\ref{sec:res:anchor} argmax traces & \path{14_argmax_alignment/} & \path{argmax_alignment.py} \\
MS MARCO set; span audit & \path{08_msmarco_counterfactual/} & \path{msmarco_tcd_audit.py}, \path{ner_span_audit.py} \\
Table~\ref{tab:arch} (ColBERT rows) & \path{07_multi_model/} & \path{colbert_on_multimodel_design.py} \\
Table~\ref{tab:arch} (other models) & \path{local/} & \path{run_multimodel_scaled.py} \\
Table~\ref{tab:regression} & \path{04_name_confound/} & \path{regression_main_grid.py}, \path{regression_rerun_projected.py} \\
Table~\ref{tab:ranking} & \path{09_ranking_impact/} & \path{realistic_ranking_impact.py} \\
Figures~\ref{fig:panel} and~\ref{fig:scale} & \path{paper/arxiv/} & \path{make_figures.py} \\
\bottomrule
\end{tabularx}
\caption{Scripts behind each reported result. Numbered directories are under \texttt{experiments/}; \texttt{local/} is at the repository root. Experiment outputs are under \texttt{results/}; corrected statistical outputs are in the dated subdirectory described in this section. Manuscript-derived summaries are under \texttt{paper/}. Dual-pipeline scripts explicitly distinguish projected and as-submitted results.}
\label{tab:repro}
\end{table}

The statistical corrections reported here are archived under \path{results/statistical_corrections_2026-11-25/}.

\paragraph{Explanatory figures.} Figures~\ref{fig:tcd} and~\ref{fig:design} are generated by \path{make_diagrams.py} in \path{paper/arxiv-2026-11-22/}. Figure~\ref{fig:tcd}c uses row 0 of the archived projected main grid: query profession doctor, template T1, Neha versus Jasmine. It is selected by file order, not by effect size. Its ten signed token differences come from \path{q1_tcd_proj.npy}; the document scores come from the \path{proj_score_a} and \path{proj_score_b} fields of \path{raw_rows_projected.json}. The plotted values were stored to six decimal places, giving a sum-to-score-gap residual of $4.94\times10^{-7}$. A small JSON extract records the values, token labels, selection rule and source-file hashes, so the diagram can be regenerated without running a model. Figure~\ref{fig:design} summarizes existing designs and adds no experimental observations.

\paragraph{Figure exports.} The two explanatory figures are supplied as vector PDF, SVG with editable text, and PNG previews. Their source data and export manifest are stored in \path{figure_sources/} beside the manuscript. The PDF includes embedded fonts, and the manuscript uses the vector versions. The manuscript source directory retains all four figure PDFs and the bibliography source.

\section{Detailed Limitations and Disclosures}
\label{app:limitations}

\paragraph{One model, one input format.} We study one checkpoint with text prefixes instead of \tok{[Q]}/\tok{[D]} markers, without query augmentation, and with special positions in the sum. The ratio and the panel comparison may differ under the official inference path; we have not checked this.

\paragraph{Small panel.} The specificity test uses 5 swap pairs per condition, and the numeral-template condition differs in wording from the identity templates. It provides no evidence that the identity ratio exceeds the numeral or invented-string ratios, but does not establish equivalence or rule out identity-related effects. A panel at main-grid scale with matched templates and a pre-specified equivalence margin is the natural next step.

\paragraph{Few token types and synthetic text.} The main grid uses one query template and short synthetic documents. Four further phrasings and two other text sources reduce but do not remove this dependence.

\paragraph{Names.} The study uses 60 US-centric first names with race and gender associations drawn from audit studies and voter-file-based data~\citep{rosenman2023}. These labels describe name associations, not individual identities. The surface-feature control does not cover frequency or semantic associations.

\paragraph{Provenance gaps.} The corrected output for the generated passages comes from a script that was not preserved. Nearly half of the MS~MARCO ``person'' spans are recogniser errors, so that set is better read as a set of arbitrary span substitutions than as a name-swap audit. The five non-ColBERT models in Table~\ref{tab:arch} were run in an earlier session, on a different profession list from the main grid. These provenance and design differences limit the corresponding comparisons.

\paragraph{No harm measure.} Nothing here measures harm to people. Ranking simulations are synthetic, and their design measures instability rather than group exposure.

\subsection{Ethical considerations}

The main controlled experiments use synthetic profiles; the MS~MARCO extension uses existing web passages with substituted spans. Race and gender categories are attached to names through published audit lists and aggregated voter-file statistics, and are used only as analytical variables; they are simplifications of social categories and say nothing about any individual. The limited evidence for specificity should not be read as evidence that retrieval models are fair to people with these names: identity swaps did move scores substantially, and ranking consequences of name changes exist. The narrower conclusion is that an elevated Func/Cont ratio alone does not establish identity-specific sensitivity.

\subsection{Acknowledgments and assistance}

We thank the anonymous reviewers of an earlier version, whose comments on the nonsense-word control, on the framing, and on ColBERT-specificity shaped this analysis.

\paragraph{Use of AI assistants.} AI coding and writing assistants were used for experiment scripts, analysis code, drafting and editing text, and figure code. Figure design and export checks drew on the scientific-visualization skill in Scientific Agent Skills~\citep{kassis2026skills}. The scoring code paths and reported numbers were checked against the implementation and stored outputs. Responsibility for the final manuscript rests with its author.

\end{document}